\pdfoutput=1

\documentclass[11pt]{article}

\usepackage[]{acl}

\usepackage{times}
\usepackage{latexsym}

\usepackage[T1]{fontenc}

\usepackage[utf8]{inputenc}

\usepackage{microtype}

\usepackage{graphicx}
\usepackage{tabularx} 
\usepackage{booktabs} 
\usepackage{mathtools}
\usepackage{siunitx}
\usepackage{amsmath}
\usepackage{amssymb}
\usepackage{euscript}
\usepackage{float}
\usepackage{caption}
\usepackage{subcaption}
\usepackage{multirow}
\usepackage{arydshln}
\usepackage{adjustbox}
\usepackage{tablefootnote}
\usepackage{todonotes}

\definecolor{tablegray}{rgb}{0.2,0.2,0.2}
\newcommand{\stdintable}[1] {~~\textcolor{tablegray}{\scriptsize{$\pm$#1}}}

\usepackage[absolute,overlay]{textpos}

\usepackage{multirow}
\usepackage{arydshln}
\usepackage{graphicx}
\usepackage{booktabs}
\usepackage{amssymb}  
\usepackage{subcaption}
\usepackage{amsmath} 

\usepackage{diagbox}
\usepackage{subcaption}

\usepackage{pifont}

\title{xGQA: Cross-Lingual Visual Question Answering}

\author{\bf Jonas Pfeiffer$^{1}$, Gregor Geigle$^{1}$, Aishwarya Kamath$^{2}$, Jan-Martin O. Steitz$^{3}$ \\
{\bf Stefan Roth$^{3}$, Ivan Vuli\'{c}$^{4}$, Iryna Gurevych$^{1}$ } \\
$^1$Ubiquitous Knowledge Processing Lab, Technical University of Darmstadt \\
 $^2$Center for Data Science, New York University \\
  $^3$Visual Inference Lab, 
  Technical University of Darmstadt \\
$^4$Language Technology Lab, University of Cambridge \hspace{0.5em} \\
}

\begin{document}
\maketitle
\begin{abstract}
 Recent advances in multimodal \textit{vision and language} modeling have predominantly focused on the English language, mostly due to the lack of multilingual multimodal datasets to steer modeling efforts. In this work, we address this gap and provide xGQA, a new multilingual evaluation benchmark for the visual question answering task. We extend the established English GQA dataset \cite{Hudson2019GQA} to 7 typologically diverse languages,  
 enabling us to detect and explore crucial challenges in cross-lingual visual question answering. We further propose new adapter-based 
 approaches   
 to adapt multimodal transformer-based models to become multilingual, and---vice versa---multilingual models to become multimodal. Our proposed 
 methods outperform current state-of-the-art multilingual multimodal models (e.g., M$^3$P) in zero-shot cross-lingual settings, but the accuracy remains low across the board; a performance drop of around 38 accuracy points in target languages showcases the difficulty of zero-shot cross-lingual transfer for this task. 
 Our results suggest that simple cross-lingual transfer of multimodal models yields latent multilingual multimodal misalignment, calling for more sophisticated methods for vision and multilingual language modeling.\footnote{The xGQA dataset is available online at: 
\url{https://github.com/Adapter-Hub/xGQA}.}
\end{abstract}


\section{Introduction}

Transformer-based architectures \cite{vaswani2017attention} have become ubiquitous in NLP \cite[][\textit{inter alia}]{Devlin2018,Liu:2019roberta, conneau2020xlmr} and 
in computer vision 
(CV)
\cite{Carion2020, Dosovitskiy2021ViT}, offering unmatched task performance. Having a shared architecture for multiple modalities opened up possibilities for effective fusion of information,  
yielding impressive performance gains across various multimodal tasks such as image captioning, phrase grounding, visual question answering, referring expression comprehension and image-text retrieval \cite[][\textit{inter alia}]{lu2019vilbert, tan2019lxmert, Li2020oscar, Zhang2020Oscarplus, Ni2021M3P, Kamath2021MDETR, Miech2021THinkFase, Frank2021Visionandlang, Bugliarello2021Multimodalumasked, Radford2021CLIP, jia2021align, Eichenberg2021Magma, Singh2021FLAVA, Fu2021VIOLET, Yang2021Crossing, Yuan2021Florence, Wang2021UFO, li2021grounded, Geigle2022Retrieve}. Yet, progress in this area has been limited mostly to the English language, as the main multimodal datasets consist only of English text. Due to the scarcity of multilingual evaluation benchmarks, there has been limited development of models that tackle this joint problem. 
 
 \begin{figure}[!t]
    \centering

        \includegraphics[width=0.95\columnwidth]{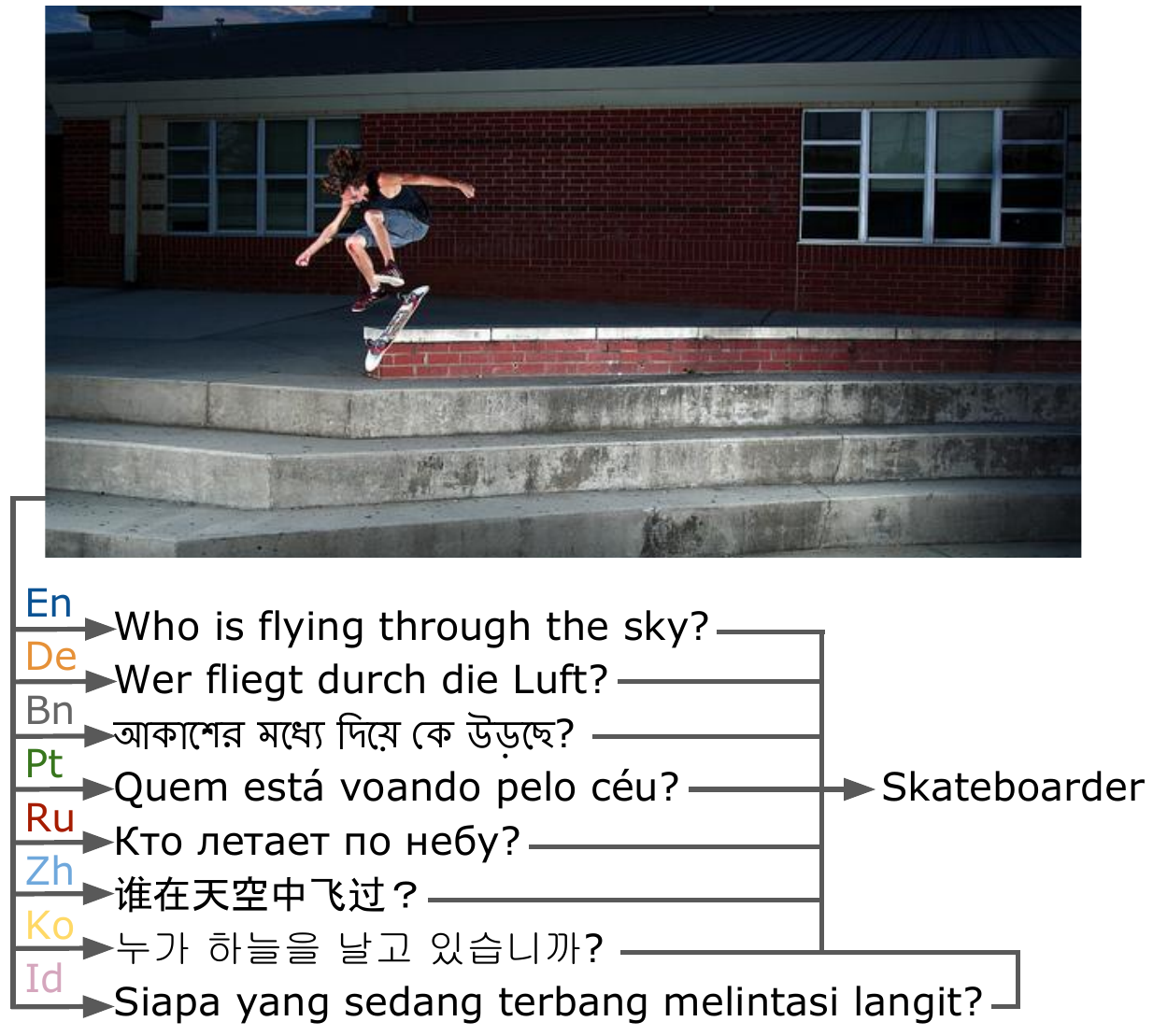}
    \caption{Example taken from the xGQA dataset with the same question uttered in 8 languages.}
    \label{fig:example}
\end{figure}

Aiming to address this gap, in this paper we propose \textbf{xGQA}, a multilingual evaluation benchmark for the visual question answering task, extending the monolingual English-only GQA dataset \cite{Hudson2019GQA}. For xGQA we manually translate and adapt the balanced GQA test-dev set into 7 new languages from 7 language families, covering 5 distinct scripts; see Figure~\ref{fig:example} and Table~\ref{tab:languages} later. In addition, we provide new fixed data splits to guide cross-lingual few-shot learning experiments, where only a small number of examples in the target language are utilized.

As pretraining is (i) notoriously computationally expensive for high-resource languages and (ii) only limited amounts of multilingual multimodal resources are available, we also propose computationally efficient adapter-based \cite{houlsby2019parameter} approaches  as additional baselines for constructing multilingual multimodal models. In a nutshell, we extend multimodal models pretrained only on English text \cite{Zhang2020Oscarplus} to become multilingual and---vice versa---multilingual models \cite{Devlin2018} to become multimodal. To this end, we follow the approaches of \citet{artetxe-etal-2020-cross} and \citet{pfeiffer-etal-2020-mad, Pfeiffer2020unks} and extend monolingual and multilingual models to new languages and scripts via learning new tokenizers and corresponding word-embedding matrices, as well as adapters for the target languages. To transfer the respective multilingual multimodal adapter-based models to the target task, we propose a novel \textit{modality-specific split architecture}, which uses modality dependent adapter weights (see Figure~\ref{fig:adapter_architecture} for an illustration of the architecture).
 
Our results clearly indicate that the proposed adapter-based architecture outperforms the recent state-of-the-art pretrained multilingual multimodal M$^3$P model \cite{Ni2021M3P} in zero-shot cross-lingual settings. However, the overall performance of zero-shot transfer remains low across the board, with an average drop of around 38 accuracy points across target languages. Using a small number of target language examples in a few-shot setup considerably improves performance for all approaches, but cross-lingual transfer performance still lags substantially behind source language performance. This demonstrates the inherent difficulty of the task, even though the corresponding questions are arguably simple as they are template based and only contain 8.5 words on average (see Figure~\ref{fig:example}).

\vspace{1.6mm}
\noindent \textbf{Contributions.} \textbf{1)}~We propose the first evaluation benchmark for cross-lingual visual question answering, covering 7 diverse target languages; \textbf{2)}~we propose novel adapter-based approaches for the creation of multilingual multimodal models; \textbf{3)}~we systematically benchmark state-of-the-art and new multilingual multimodal models in zero-shot and few-shot learning setups, demonstrating the difficulty of the proposed task and serving as strong reference points for future work; \textbf{4)}~we provide a thorough analysis of the different approaches, highlighting the aspects and question types that lead to the most common model failures, again motivating future work in this domain.
 
\section{Background and Related Work}

\noindent\textbf{Multilingual Language Models.}
Pretrained multilingual transformer-based LMs such as mBERT \cite{Devlin2018} and XLM-R \cite{conneau2020xlmr} adopt the same pretraining regime as their respective monolingual counterparts: BERT \cite{Devlin2018} and RoBERTa \cite{Liu:2019roberta}. They are pretrained via self-supervised masked language modelling objective (MLM) on concatenated text corpora of more than 100 languages, where text is tokenized using WordPiece, SentencePiece or BytePair encodings. These multilingual models have been shown to work surprisingly well for cross-lingual tasks, despite the fact that they do not rely on  direct cross-lingual supervision  \cite[e.g., parallel data, translation dictionaries;][]{pires:2019, wu-dredze-2019-beto,artetxe-etal-2020-cross,Hu2020xtreme, k:2020, Rust2021tokenizer}.

\vspace{1.6mm}
\noindent\textbf{Vision and Language Models.}
Most transformer-based multimodal models \cite[][\textit{inter alia}]{lu2019vilbert,tan2019lxmert,Chen2019,Li2019,gan2020large, Li2020oscar,Bugliarello2021Multimodalumasked,Ni2021M3P} jointly encode text tokens and image region features by preprocessing images using object detection models---such as Faster R-CNN \cite{ren2015faster}---to extract features for regions of interest (RoI) \cite{butd-anderson-2018}.
The image region features are passed through an affine layer, which learns to project the region features to the joint embedding space of the multimodal transformer. 
The bounding box coordinates of the RoI act as positional embeddings for the visual features. As such, they  undergo an affine transformation to the embedding space and are combined with their respective image region representation. The position-aware image region embeddings get passed into the transformer.
The multi-head attention then attends over all text and image inputs at every layer, learning a joint representation of both modalities. On the other hand, \citet{Kamath2021MDETR} avoid using object detectors as a black-box for pre-extracting these region features and instead make it a central part of the multimodal transformer architecture. Training the object detector end-to-end with the multimodal transformer 
adds 
flexibility and better representation capacity.

Similar to MLM,  
multimodal transformer-based models are trained with self-supervised objectives such as masked feature regression, masked object detection, masked attribute detection, and contrastive losses such as cross-modality matching \cite{tan2019lxmert}. 
Typically, image captioning datasets are used for pretraining such as COCO \cite{lin2014microsoft}, Flickr30k \cite{Plummer2015Flickr}, Conceptual Captions (CC) \cite{sharma-etal-2018-conceptual}, and SBU \cite{Ordonez:2011:im2text}. Similar to unimodal language models, the [CLS] token is  used as a contextual representation for classification tasks.  

Multilingual multimodal models have also been proposed recently: M$^3$P \cite{Ni2021M3P} is trained on the Wikipedias of 50 different languages and the English multimodal CC dataset. In order to align tokens of languages other than English with image representations, M$^3$P utilizes a code-switching mechanism, where words of the English CC examples are randomly replaced with words from corresponding bilingual dictionaries. 
In UC$^2$, \citet{Zhou2021UC2} augment English multimodal datasets with other languages via machine translation and propose masked region-to-token modeling and visual translation language modeling.\footnote{The model weights of UC$^2$  were not released by the time of experimentation.}

\vspace{1.6mm}
\noindent\textbf{Adapters} \cite{Rebuffi2017adapters,houlsby2019parameter} have been introduced as a more efficient fine-tuning strategy for transfer learning in NLP and CV. Instead of fine-tuning all the weights of a pretrained model on the target task, small feed-forward layers are introduced at each layer of the pretrained model. During task fine-tuning, only the adapter weights are updated, while the pretrained parameters remain fixed/frozen.
Adapters have been shown to be very training efficient \cite{rueckle2020adapterdrop}, and among an increasing amount of applications they can be utilized to transfer between domains \cite{ruckle-etal-2020-multicqa} and tasks \cite{Poth2021Pretrain}, and in  machine translation~\citep{bapna-firat-2019-simple, philip-etal-2020-monolingual, Le:2021acl} and cross-lingual transfer~\cite[\textit{inter alia}]{pfeiffer-etal-2020-mad, Pfeiffer2020unks, ustun-etal-2020-udapter,ansell-etal-2021-mad-g} scenarios.  

\vspace{0.5em}
\noindent\textbf{Datasets.} 
Pretraining and fine-tuning data for multilingual multimodal models is typically based on (multimodal information from) Wikipedia 
 \cite[\textbf{WikiCaps}, \textbf{WIT},][]{Schamoni2018WikiCaps,Srinivasan2021WIT}, or on available downstream task data.
\textbf{Multi30k} \cite{elliott-etal-2016-multi30k} is a multilingual image captioning dataset for retrieval-type questions, covering English, German, French, and Czech; 
\textbf{GEM} \cite{su2021GEM} covers image and video retrieval tasks across 20 and 30 different languages, respectively; \textbf{HowTo100M} \cite{huang-etal-2021-multilingual} is a multilingual and multimodal pretraining dataset for image and video retrieval; 
\textbf{MultiSubs} \cite{Wang2021MultiSubs} focuses on fill-in-the-blank tasks and lexical translation, covering English, Spanish, German, Portuguese, and French. 
\citet{Gao2015Talkingtoamachine, Shimizu2018BiVQA} propose bilingual visual question answering datasets for English, and Chinese and Japanese respectively.
In contemporary work \citet{Liu2021VisGroundCultures} propose \textbf{MaRVL}, a binary multilingual question answering dataset similar to NLVR2 \cite{suhr-etal-2019-corpus}, spanning 5 typologically diverse languages (Chinese, Tamil, Swahili, Indonesian, and Turkish). 

Previous datasets predominantly focus on (arguably simpler) retrieval-type tasks, only cover a small set of similar languages (e.g., Multi30k, MultiSubs), or only cover binary questions. In contrast, we propose the first multilingual visual question answering dataset, which covers a typologically more diverse set of languages. 

Most recently, \textbf{IGLUE} \cite{Bugliarello2022IGLUE}---a multilingual multimodal benchmark that integrates xGQA---was proposed: IGLUE brings together visual question answering, cross-modal retrieval, grounded reasoning, and grounded entailment tasks across 20 diverse languages. 

\section{xGQA}
\label{sec:xGQAData}
The original English GQA dataset \cite{Hudson2019GQA} was constructed by leveraging Visual Genome scene graphs \cite{Krishna2017VG}. An English question engine that utilizes \textit{content} (i.e.~information about objects, attributes, and relations provided) and \textit{structure} (a linguistic grammar that couples hundreds of structural patterns and detailed lexical semantic resources) was used to generate over 22 million diverse questions, which are visually grounded in the image scene graphs. As the questions are automatically generated using templates, they do not necessarily reflect the wide spectrum of natural language, making any assumptions on the performance in the wild difficult. 

Each question is associated with additional metadata such as \textbf{structural types}: 
(1) \textit{verify} for yes/no questions (e.g. "Do you see any cats?"), (2) \textit{query} for all open questions (e.g. "Who is wearing jeans?"), (3) \textit{choose} for questions that present two alternatives
to choose from (e.g. “Is it red or blue?”), (4) \textit{logical} which
involve logical inference (e.g. "Is the field soft and snowy"), and (5) \textit{compare} for comparison
questions between two or more objects (e.g. "Are all the animals zebras?"). For further details regarding the metadata, we refer the reader to \citet{Hudson2019GQA}. 

\vspace{1.6mm}
\noindent\textbf{Dataset Design.} The principal objective when devising xGQA was to create a genuinely typologically diverse multimodal and multilingual evaluation benchmark for visual question answering. We utilize the balanced\footnote{To reduce biases in the conditional answer distribution \citet{Hudson2019GQA} utilize the structural metadata to downsample and create balanced datasets that are more robust against shortcuts and
guesses.} test-dev set of GQA, which consists of 12,578 questions about 398 images.\footnote{We chose to translate the test-dev set of GQA, as the labels for test-std are not released. } Due to the defined structural patterns, the formulation of the questions is simple, with an average length of 8.5 words.\footnote{For this reason, we chose to hire  university students that are currently conducting their (Computer Science or Computational Linguistics) studies in English and are all fluent English speakers to translate the question into their native language. They were paid above the minimum hourly wage of the country of their respective university. After all questions have been translated, another, independent native speaker then verified the translations based on random spot checks.} The resulting xGQA dataset covers translations in 7 languages, each representing a distinct language family, and contains examples written in 5 different scripts (see Table~\ref{tab:languages}). 

\begin{table}[!t]
\centering
\footnotesize
\def\arraystretch{1.0}
\resizebox{\columnwidth}{!}{
\begin{tabular}{llllcc}
\toprule
Language & iso & Family & Script & Speakers \\
\midrule
English & en & IE:Germanic & Latin & 400M \\
German & de & IE:Germanic & Latin & 95M \\
Portuguese & pt & IE:Romance & Latin & 250M \\
Russian & ru & IE:Slavic & Cyrillic & 150M \\
Indonesian & id & Austronesian & Latin & 43M \\
Bengali & bn & IE:Iranian & Bengali & 230M \\
Korean & ko & Koreanic & Korean & 77M \\
Chinese & zh & Sino-Tibetan & Chinese & 1.2B \\
\bottomrule
\end{tabular}
}
\caption{Languages covered by xGQA. IE stands for Indo-European.}  
\label{tab:languages}
\end{table}

\vspace{1.6mm}
\noindent\textbf{Few-Shot Data Splits.} In order to conduct cross-lingual  few-shot learning experiments, we provide new data splits of different sizes. We split on images and add all questions associated with the image to the respective set. The development and test sets consist of 50 and 300 images, respectively. The training splits consist of 1, 5, 10, 20, 25, and 48 images, see Table~\ref{tab:datasetsizes}. We ensure that the distribution of structural types within each set is maintained. 

xGQA is the first truly typologically diverse multilingual multimodal benchmark, unlocking new experimentation and analysis opportunities in cross-lingual zero-shot and few-shot scenarios. While the questions in xGQA are intuitive and easy for humans to solve, we later show that current state-of-the-art models still have difficulty with  transfer.

\begin{table}[]
\centering
\footnotesize
\def\arraystretch{1.0}
\resizebox{\columnwidth}{!}{
\begin{tabular}{lrrrrrrrr}
 \toprule
Set         & \multicolumn{1}{c}{Test} & \multicolumn{1}{c}{Dev} & \multicolumn{6}{c}{Train}         \\
\cmidrule(lr){2-2} \cmidrule(lr){3-3} \cmidrule(lr){4-9}
\#Img    & 300                      & 50                      & 1  & 5   & 10  & 20  & 25  & 48   \\
\#Ques & 9666                     & 1422                    & 27 & 155 & 317 & 594 & 704 & 1490 \\
\bottomrule
\end{tabular}
}
\caption{Few-shot dataset sizes. The GQA test-dev set is split into new development, test sets, and training splits of different sizes. We maintain the distribution of structural types in each split.}
\label{tab:datasetsizes}
\end{table}

\section{Baselines}
 
To analyze the performance and current gaps on xGQA, we first evaluate the recently proposed M$^3$P model, which has been pretrained on multilingual and multimodal data. However, pretraining is computationally expensive and only limited amounts of multilingual multimodal resources are available. Therefore, we further propose new and more efficient approaches that (1) extend state-of-the-art multilingual language models to the multimodal domain and (2) provide multilingual capabilities to state-of-the-art multimodal models.
 
Unless noted otherwise, we follow the predominant fine-tuning strategy for  GQA;  a prediction head is placed on top of the output of a pretrained transformer. All possible 1853 answers of the GQA task are mapped to a class label.  The question associated with an image together with the position-aware region features are passed as input to the transformer, supervised using a cross-entropy loss.\footnote{For instance, we use this strategy to fine-tune all parameters of M$^3$P on the GQA training data.}

 \begin{figure}[!t]
    \centering
        \includegraphics[width=0.75\columnwidth]{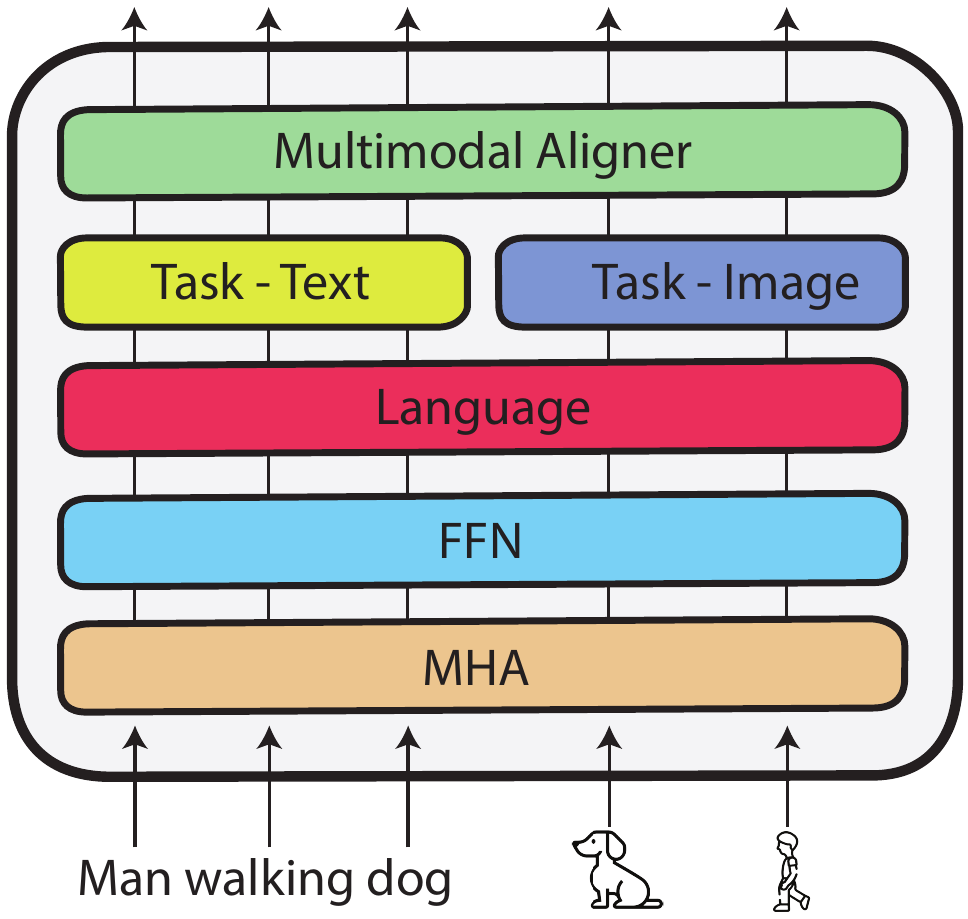}
    \caption{Architecture of an adapter-based multilingual multimodal model. Text and image inputs share the weights of the multi-head attention (MHA) and feed-forward (FFN) layers, as well as the \textit{language} and \textit{multimodal align} adapters. Each modality is passed through a modality specific \textit{task} adapter, the outputs of which are concatenated.  }
    \label{fig:adapter_architecture}
\end{figure}

\subsection{Multimodal $\rightarrow$ Multilingual}
\label{sec:multimodal_to_multilingual}
\noindent\textbf{OSCAR+$^{Emb}$.} To extend a monolingual transformer LM to a multilingual domain, \citet{artetxe-etal-2020-cross}  fine-tune a new word-embedding layer in the target language.  
Inspired by this idea, we now describe how we extend the current state-of-the-art monolingual multimodal transformer model \text{OSCAR+}~\cite{ Zhang2020Oscarplus} to learn new embeddings for the target languages. 

In the \textit{language-extension} phase, we  replace the embedding matrix of OSCAR+ with a randomly initialized embedding matrix.\footnote{Following \citet{Pfeiffer2020unks}, we copy the embeddings of lexically overlapping tokens (if such tokens exist) from the original embedding space to the new embedding space, as it typically works better than fully random initialization.} The transformer weights are frozen while only the newly introduced embeddings are fine-tuned on unlabeled text data of the target language with the MLM objective.

In the \textit{target-task} phase, the original OSCAR+ model is fine-tuned on the English training data of GQA, where the transformer layers are fine-tuned, but the embedding layer is frozen. During inference, the embedding layer is replaced with the target language's embedding layer. 

\vspace{1.6mm}
\noindent\textbf{OSCAR+$^{Ada}$.} We  extend this by adding adapters.   

In the \textit{language-extension} phase  we follow \citet{Pfeiffer2020unks} in order to extend the model to the target languages. Similar to OSCAR+$^{Emb}$, we train a new embedding layer. 
We further add  \textit{language} adapters at every transformer layer. 
Given that  OSCAR+ is trained on English text, we follow  \citet{pfeiffer-etal-2020-mad} when training English \textit{language} adapter modules, without replacing the embedding matrix. The transformer weights are frozen while only the \textit{newly} introduced embeddings and  \textit{language} adapter weights are fine-tuned on unlabeled text data of the language. 

For the \textit{target-task} phase, we propose a novel modality-split architecture (see Figure~\ref{fig:adapter_architecture}) inspired by the cross-lingual transfer method of \citet{pfeiffer-etal-2020-mad}. At each transformer layer, text and image representations are passed through the pretrained  multi-head attention (MHA) and feed-forward (FFN) layers. Both image and text representations are also passed through the pre-trained \textit{language} adapters. Each modality is then passed through modality-specific \textit{text} and \textit{image} \textit{task} adapters and next through a shared \textit{multimodal alignment} adapter.\footnote{We have compared multiple different architectures as illustrated in Figure~\ref{fig:alternativ_ada_settings} in the Appendix, finding this setup to perform best. We present results of the alternative architectures also in the Appendix.} 
We follow \citet{pfeiffer-etal-2020-mad}, freezing transformer, embedding and \textit{language} adapter weights during training, thus fine-tuning only the \textit{task} and \textit{multimodal aligner} adapter weights, together with the prediction head. At inference time, the embedding layer and the \textit{language} adapters are replaced with the target language weights. 

\subsection{Multilingual $\rightarrow$  Multimodal}
\noindent\textbf{mBERT$^{Ada}$.}
For experiments where we extend a multilingual model to become multimodal, we utilize mBERT \cite{Devlin2018}.

Given that mBERT is able to represent many different languages, it is not necessary to learn new embedding layers for the target languages in the \textit{language-extension} phase. Instead, we utilize the mBERT-compatible \textit{language} adapters available on \href{https://AdapterHub.ml}{AdapterHub.ml} \cite{pfeiffer-etal-2020-adapterhub}.\footnote{While all xGQA languages already have readily available language adapters on AdapterHub, any hypothetical extension of experiments to languages without such adapters would involve training their dedicated language adapters, e.g., following the procedure of \newcite{pfeiffer-etal-2020-mad}.}

For the \textit{target-task} phase, we follow OSCAR+ for the image representation layer, where image features are combined with their respective positional information and passed through an affine transformation layer.  We experiment with the same adapter architecture from Figure~\ref{fig:adapter_architecture}, as described for  OSCAR+$^{Ada}$. We again freeze transformer, embedding and \textit{language} adapter weights during training. However, in contrast to OSCAR+$^{*}$, we randomly initialize and fine-tune the affine image transformation layer. We also fine-tune the \textit{task}, \textit{multimodal aligner} adapter weights, and prediction head, all on the GQA task. At inference time, the embedding layer and the \textit{language} adapters are replaced with the corresponding target language weights.

\section{Experimental Setup}

\subsection{Language-Extension Phase}
For OSCAR+$^{Emb}$ and OSCAR+$^{Ada}$, we follow the general setups proposed by \citet{pfeiffer-etal-2020-mad,Pfeiffer2020unks}. We train a new word-piece tokenizer for each target language with a vocabulary size of 30k. We fine-tune the randomly initialized embedding layer, and (for OSCAR+$^{Ada}$) adapter layers for 100k update steps with a batch size of 64 and a learning rate of $1\mathrm{e}{-4}$. For mBERT$^{Ada}$, we utilize the language adapters from \href{https://AdapterHub.ml}{AdapterHub.ml}. 

\subsection{Fine-tuning on GQA}
\label{sec:exp:finetuning}
We follow the standard setup proposed by \citet{Li2020oscar}, passing the representation of the [CLS] token through a prediction head. We fine-tune the respective models using a cross-entropy loss with labels being all possible answers in the GQA dataset. Following prior work \cite{Li2020oscar}, we use a batch size of 192 and train for 5 epochs on the unbalanced GQA training portion.

\vspace{1.2mm}
\noindent\textbf{M$^3$P.} We fine-tune all weights of the pretrained model with a learning rate of $3\mathrm{e}{-5}$.

\begin{table*}[t!]
\centering
\resizebox{1.0\textwidth}{!}{%
\begin{tabular}{ll:lllllll|l}
\toprule
model &      en &     de &     pt &     ru &     id &     bn &     ko &     zh &   mean   \\
\midrule
M3P & 58.43\stdintable{1.4} &	23.93\stdintable{3.2}	& 24.37\stdintable{4.0}	& 20.37\stdintable{3.4} & \bf	22.57\stdintable{6.1} & \bf	15.83\stdintable{3.6} &	16.90\stdintable{3.8} &	18.60\stdintable{1.0} &	20.37 \\
OSCAR+$^{\text{Emb}}$ &   \bf 62.23\stdintable{0.3} &  17.35\stdintable{1.0} &  19.25\stdintable{0.4} &  10.52\stdintable{4.0} &  18.26\stdintable{0.4} &  14.93\stdintable{2.0} &  17.10\stdintable{1.8} &  16.41\stdintable{3.2} &  16.26 \\ 
OSCAR+$^{\text{Ada}}$ &   60.30\stdintable{0.4} &  18.91\stdintable{0.8} &  27.02\stdintable{2.3} &  17.50\stdintable{1.2} &  18.77\stdintable{0.3} &   15.42\stdintable{2.0} &  15.28\stdintable{2.7} &  14.96\stdintable{2.1} &  18.27 \\
mBERT$^{\text{Ada}}$ &   56.25\stdintable{0.5} & \bf   29.76\stdintable{2.3} & \bf  30.37\stdintable{1.8} & \bf  24.42\stdintable{1.1} &  19.15\stdintable{2.8} &  15.12\stdintable{1.9} & \bf  19.09\stdintable{0.9} & \bf  24.86\stdintable{1.8} &  \bf 23.25 \\
\bottomrule
\end{tabular}
}
\caption{Zero-shot transfer results when transferring from English GQA. Average accuracy and standard deviation are reported. Best results  are highlighted in \textbf{bold}; \textit{mean} scores are not averaged over the source language (English).}
\label{tab:new_models_results}
\end{table*}

\noindent\textbf{OSCAR+$^{Emb}$, OSCAR+$^{Ada}$, and mBERT$^{Ada}$.} We use the pretrained weights and image region features provided by \citet{Zhang2020Oscarplus}. However, we do not pass the object attribute labels as inputs to the model. The object attribute labels are in English and utilizing them in cross-lingual scenarios is non-trivial.\footnote{The replaced tokenizer and embedding representations of the target language potentially do not adequately represent English terms, resulting in a misalignment between the question (in the target language) and the object attributes (in English). } We leave this for future work.

For the OSCAR+$^{Emb}$ setting, we fine-tune the transformer weights and the prediction head and freeze the embedding layer, using a learning rate of $3\mathrm{e}{-5}$.
For the OSCAR+$^{Ada}$ and mBERT$^{Ada}$ settings, we add adapter layers as described in \S\ref{sec:multimodal_to_multilingual} and illustrated in Figure~\ref{fig:adapter_architecture}. We freeze all pretrained weights--including embeddings, transformer layers, and language adapters--and only fine-tune the newly introduced adapters and the prediction head. For mBERT$^{Ada}$, we also add and train the affine image transformation layer. We fine-tune  the adapter-based models with a learning rate of $1\mathrm{e}{-4}$.

\subsection{Zero-Shot Cross-Lingual Transfer}
\label{sec:exp:zeroshot}

For zero-shot cross-lingual evaluation, we utilize the model fine-tuned on the GQA training data and evaluate on the multilingual xGQA test data.  
The model checkpoint that performed best on the English GQA validation data is selected for transfer. 

\vspace{1.1mm}
\noindent\textbf{M$^3$P.} As the model is pre-trained to cover, among others, xGQA languages, no additional steps are required for cross-lingual transfer. 

\vspace{1.1mm}
\noindent\textbf{OSCAR+$^{Emb}$.} We replace the English embedding layer with the target-language embedding layer.

\vspace{1.1mm}
\noindent\textbf{OSCAR+$^{Ada}$.} We replace the English embedding and language adapter layers with the embedding and adapters layers of the target language.

\vspace{1.1mm}
\noindent\textbf{mBERT$^{Ada}$.}  We replace the language adapter layers with the  adapters layers of the target language.

\subsection{Few-Shot Cross-Lingual Transfer}
\label{sec:exp:fewshot}
For few-shot cross-lingual scenarios we follow \citet{Lauscher:2020zerohero} and start from the same fine-tuned model as for zero-shot transfer (see \S\ref{sec:exp:zeroshot}). We then fine-tune the same parts of the model as when training on the English training data as in \S\ref{sec:exp:finetuning}, but on the small portions of multimodal data available in the target language. We train on the different data splits, consisting of 1, 5, 10, 15, 20, 25, and 48 images (see Table~\ref{tab:datasetsizes}). We experiment with training for a different number of epochs (5, 10) using different learning rates ($1\mathrm{e}{-5}$ and $5\mathrm{e}{-5}$ for M$^3$P and OSCAR+$^{Emb}$, and $5\mathrm{e}{-5}$ and $1\mathrm{e}{-4}$ for OSCAR+$^{Ada}$ and mBERT$^{Ada}$). We find that training for longer and with a larger learning rate performed best for all settings.

\section{Results and Discussion}
The main results are presented in Table~\ref{tab:new_models_results} (zero-shot experiments) and in Table~\ref{tab:fewshot_results} (few-shot).

\subsection{Zero-Shot Cross-Lingual Transfer}
\label{ss:zero_results}
One of our core findings is that multimodal zero-shot cross-lingual transfer is extremely difficult; we witness an average drop in accuracy  of more than 38 points on the target languages of the xGQA dataset compared to English GQA scores (e.g., compare the results with M$^3$P).  

While, as expected, OSCAR+ achieves the best accuracy on the English test set, the massively multilingual models---M$^3$P and mBERT---perform considerably better in cross-lingual transfer.\footnote{The superior accuracy of OSCAR+ on the English test set is expected as the model was pretrained on large English multimodal data. 
We find that fine-tuning all transformer weights (OSCAR+$^{Emb}$)  achieves slightly better results than only training adapter weights (OSCAR+$^{Ada}$). Our slightly lower scores compared to results by \citet{Zhang2020Oscarplus} can be explained by us (1) not fine-tuning the embedding layer, and (2) not utilizing the attribute labels. Further, previous works that focus only on English add the official \textit{validation} set to the \textit{training} set, use the official \textit{test-dev} set as their dev set, and report their test scores of the official GQA test benchmark \textit{test-std} for which labels are not available.  Our scores follow the training splits, where we use the official \textit{test-dev} set as the final test set, as described before in \S\ref{sec:xGQAData}.}  This indicates, that joint multilingual pretraining is important and a simple multilingual adapter-based or embedding-based extension of monolingual models achieves inferior cross-lingual performance. 

While the pretraining method M$^3$P achieves better accuracy on the English test set, the adapter-based multimodal extension of mBERT outperforms M$^3$P in cross-lingual transfer. We hypothesize that, when fine-tuning all transformer weights on monolingual multimodal data, the cross-lingual alignment breaks within M$^3$P. However, this does not happen in adapter-based settings, as the multilingual weights are frozen and thus remain intact.

\begin{figure}[!t]
    \centering
    \begin{subfigure}[b]{\columnwidth}
        \centering
        \includegraphics[width=\columnwidth]{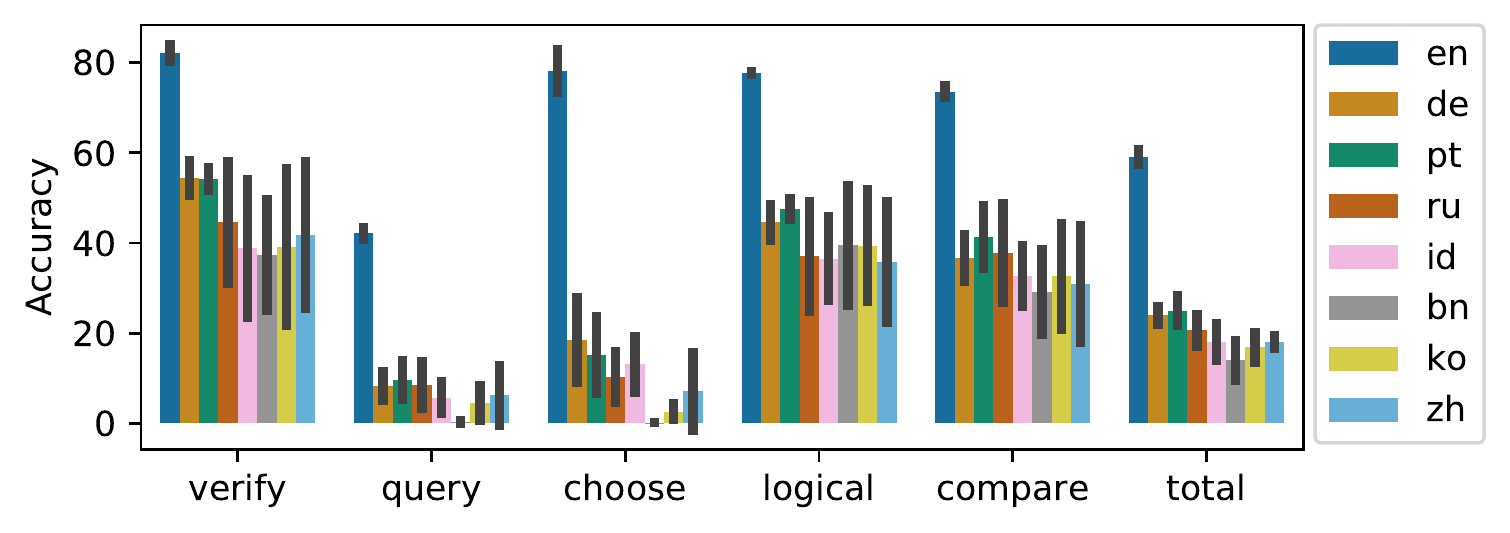}
        \caption{M3P}
        \label{fig:zerom3p}
    \end{subfigure}
    \hspace{-1em}
    \begin{subfigure}[b]{\columnwidth}
        \centering
        \includegraphics[width=\columnwidth]{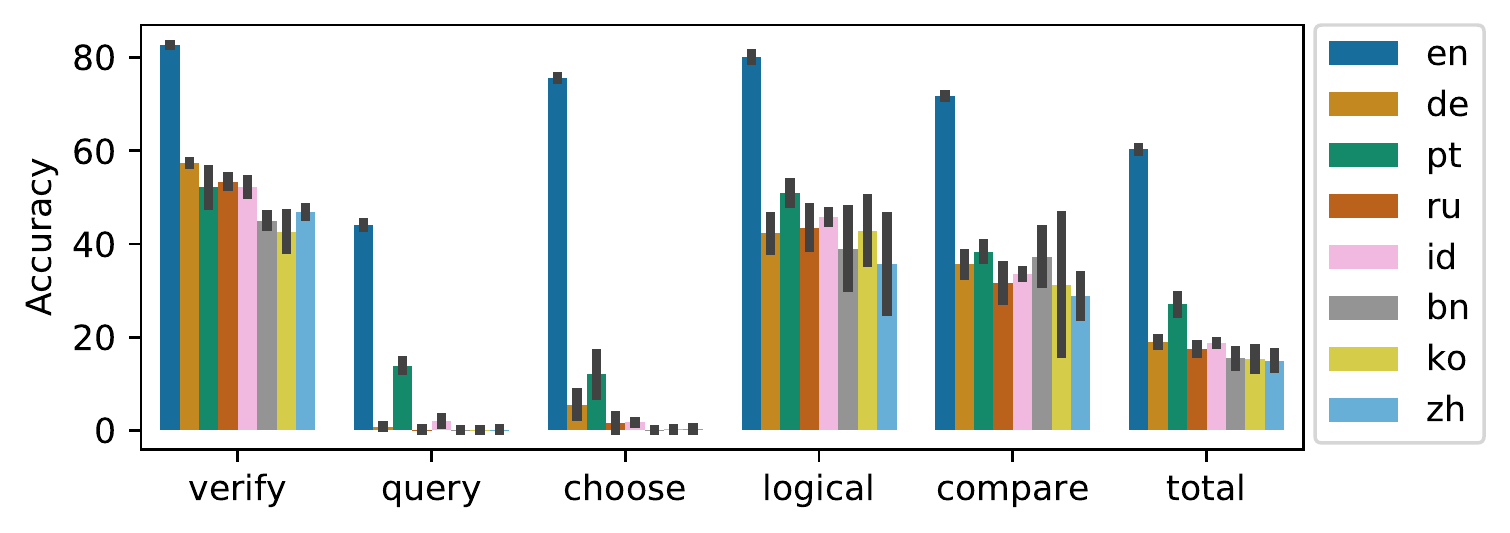}
        \caption{OSCAR+$^{\text{Ada}}$}
        \label{fig:zerombert}
    \end{subfigure}
    \hspace{-0.5em}
    \begin{subfigure}[b]{\columnwidth}
        \centering
        \includegraphics[width=\columnwidth]{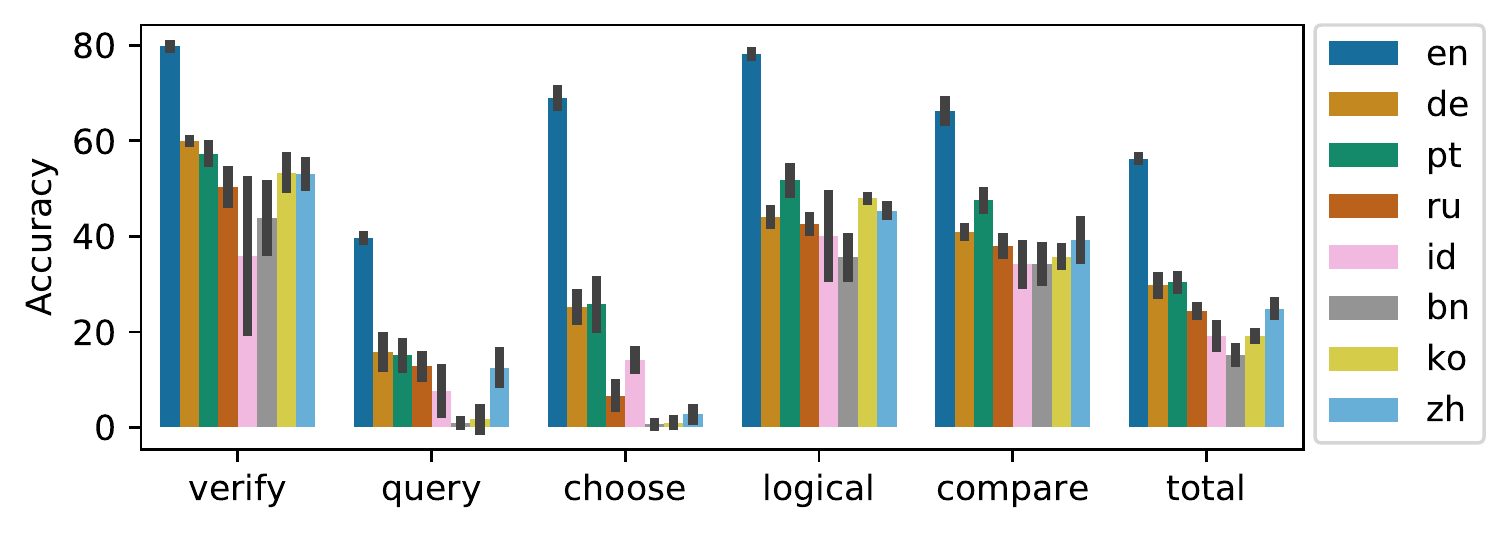}
        \caption{mBERT$^{\text{Ada}}$}
        \label{fig:zerooscar} 
    \end{subfigure}
    \caption{Zero-shot accuracy across different languages and structural question types from xGQA.}
    \label{fig:zerostructural}
\end{figure}

\vspace{1.6mm}
\noindent \textbf{Analysis of Structural Question Types.}  
Figure~\ref{fig:zerostructural} depicts our analysis of the structural question types in  zero-shot experiments. We observe large drops in accuracy especially for \textit{query} and \textit{choose} type questions.  \textit{Query} type questions are free-form  and thus semantically the most difficult  to answer, even in the source language (English). This explains the overall low accuracy across all approaches in zero-shot settings for this question type.

\begin{figure}[!t]
    \centering
    \begin{subfigure}[b]{\columnwidth}
        \centering
        \includegraphics[width=\columnwidth]{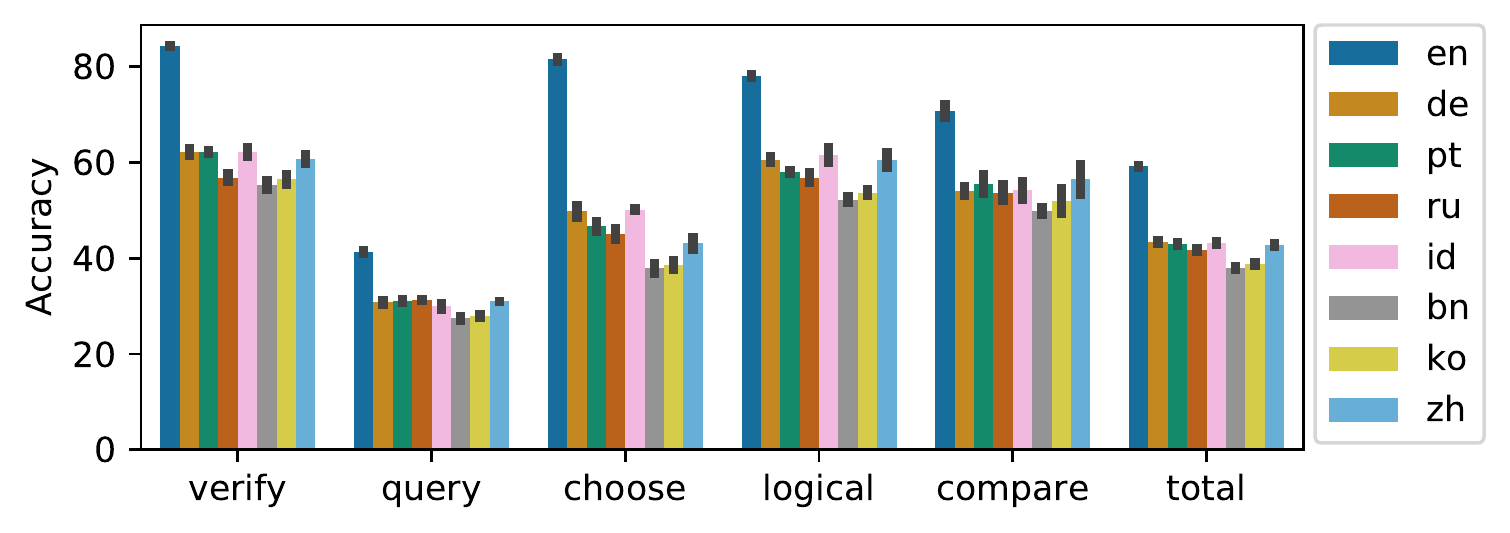}
        \caption{M3P}
        \label{fig:zerom3p}
    \end{subfigure}
    \hspace{-1em}
    \begin{subfigure}[b]{\columnwidth}
        \centering
        \includegraphics[width=\columnwidth]{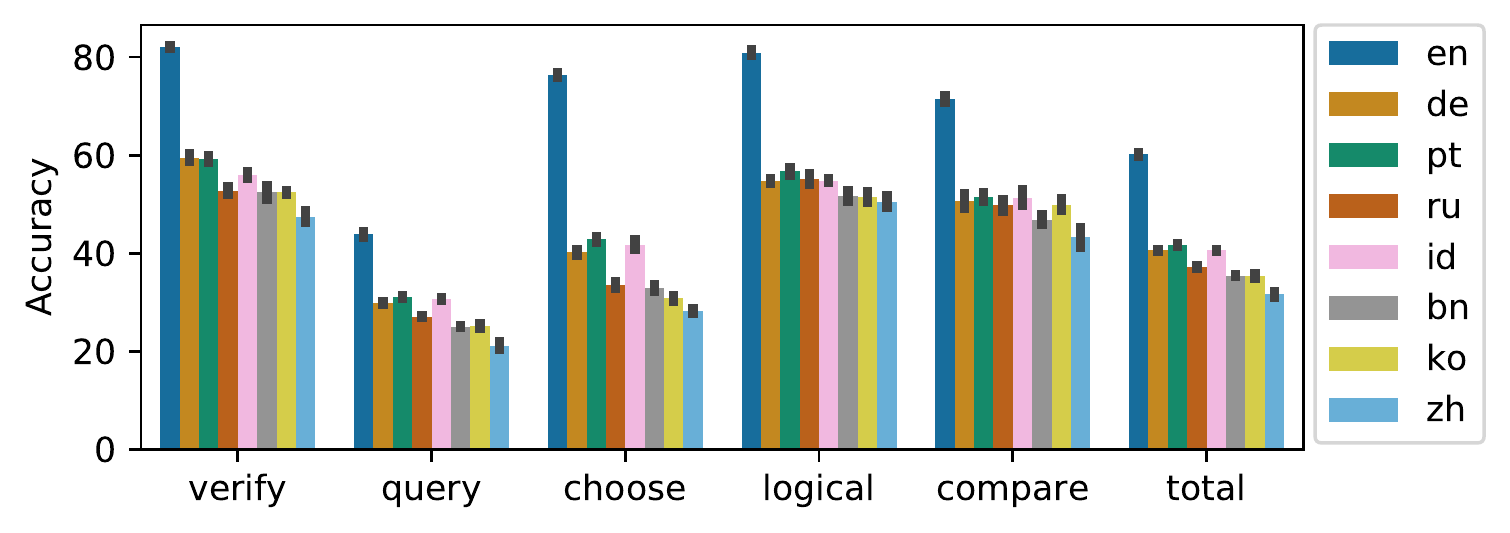}
        \caption{OSCAR+$^{\text{Ada}}$}
        \label{fig:zerombert}
    \end{subfigure}
    \hspace{-0.5em}
    \begin{subfigure}[b]{\columnwidth}
        \centering
        \includegraphics[width=\columnwidth]{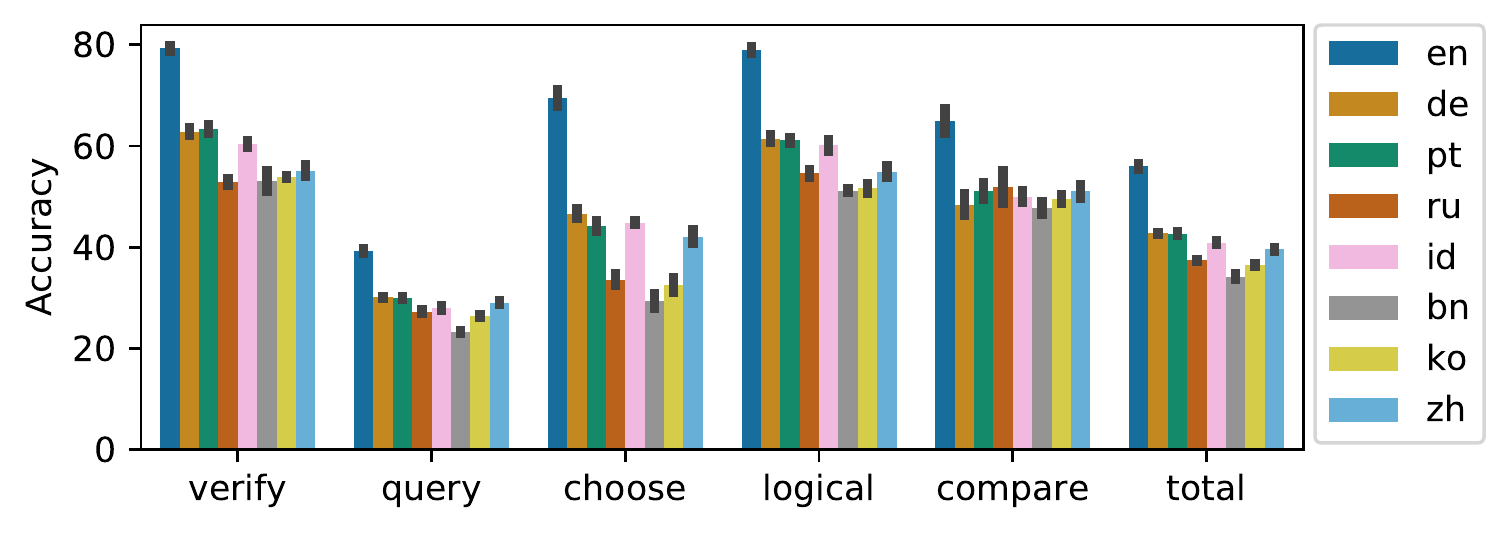}
        \caption{mBERT$^{\text{Ada}}$}
        \label{fig:zerooscar}
    \end{subfigure}
    \caption{Few-shot accuracy (with 48 images) across different languages and  question types from xGQA.}
    \label{fig:fewshotstructural}
\end{figure}

This is in stark contrast with the  \textit{choose}-type questions, which the models perform very well on in the source language. However, we report a substantial accuracy drop in zero-shot cross-lingual transfer. This decrease is most likely due to the nature of the question formulation and the modelling implementation. \textit{Choose}-type questions are formulated such that the answer to the question is a word or phrase which appears in the question, i.e. "Is it \underline{red} or \underline{blue}?". The label classes, and consequently the prediction head, are constructed as a set of all answers appearing in the dataset. This means that the model learns a distributed representation of each answer in its final layer. Consequently, in cross-lingual transfer, 
the model is required to automatically align the question's options "\underline{red}" or "\underline{blue}" (translated in their respective language), with their English latent representation of the  model's prediction head. The very low results in this category  indicate that this cross-lingual word alignment breaks in zero-shot scenarios.

Overall, zero-shot transfer with our proposed multimodal adapter-based extension of mBERT (mBERT$^{Ada}$) achieves the best accuracy, with almost 3 points increase over M$^3$P and almost 5 points increase over OSCAR+. However, the overall accuracy of all approaches remains low in comparison to the results in English. This indicates that zero-shot multimodal cross-lingual transfer is extremely difficult, most likely due to the misalignment issue between visual and cross-lingual internal representations. To investigate this conjecture further, we run similar tests in few-shot setups, which should potentially mitigate the misalignment issue observed in zero-shot setups.

\begin{table}[t!]
    \centering
    \def\arraystretch{0.93}
\resizebox{1.0\columnwidth}{!}{%
\begin{tabular}{llrrrrrrr}
\toprule
\multirow{2}{*}{\textbf{Lang}} & \multirow{2}{*}{\textbf{Model}}  & \multicolumn{7}{c}{\bf \# Training Images}\\
 & &  \multicolumn{1}{c}{\textbf{0}}  &   \multicolumn{1}{c}{\bf  1 } & \multicolumn{1}{c}{  \bf  5}  & \multicolumn{1}{c}{   \bf 10 } &\multicolumn{1}{c}{  \bf   20} &  \multicolumn{1}{c}{ \bf  25} & \multicolumn{1}{c}{  \bf  48 }  \\
\midrule
\multirow{4}{*}{de} & M3P &  24.78 &  31.49 &\bf39.31 &\bf41.05 &\bf42.22 &\bf42.54 &\bf43.16 \\
   & OSCAR+$^{\text{Emb}}$ &  17.49 &  17.84 &  29.09 &  34.48 &  37.35 &  38.45 &  41.08 \\
   & OSCAR+$^{\text{Ada}}$ &  17.84 &  21.40 &  31.26 &  35.84 &  37.92 &  38.46 &  40.58 \\
   & mBERT$^{\text{Ada}}$ &\bf32.41 &\bf33.87 &  37.44 &  39.15 &  40.65 &  41.63 &  42.71 \\
      \midrule
\multirow{4}{*}{pt} & M3P &  26.73 &  32.98 &  37.23 &\bf39.07 &\bf40.92 &\bf41.05 &  43.06 \\
   & OSCAR+$^{\text{Emb}}$ &  19.36 &  22.55 &  32.42 &  36.37 &  39.01 &  40.15 &\bf43.27 \\
   & OSCAR+$^{\text{Ada}}$ &  24.58 &  29.61 &  34.73 &  37.46 &  38.82 &  39.70 &  41.75 \\
   & mBERT$^{\text{Ada}}$ &\bf31.45 &\bf33.27 &\bf37.31 &  38.88 &  40.51 &  41.03 &  42.62 \\
     \midrule
\multirow{4}{*}{ru} & M3P &  24.29 &\bf32.32 &\bf36.71 &\bf38.53 &\bf39.94 &\bf40.13 &\bf41.85 \\
   & OSCAR+$^{\text{Emb}}$ &   7.98 &  17.32 &  23.72 &  28.21 &  32.15 &  32.87 &  36.84 \\
   & OSCAR+$^{\text{Ada}}$ &  16.38 &  19.74 &  27.42 &  30.17 &  33.22 &  34.21 &  37.28 \\
   & mBERT$^{\text{Ada}}$ &\bf25.51 &  26.47 &  31.69 &  32.47 &  34.93 &  35.53 &  37.42 \\
     \midrule    
\multirow{4}{*}{id} & M3P &  18.74 &  31.37 &\bf37.24 &\bf38.65 &\bf41.07 &\bf42.00 &\bf43.12 \\
   & OSCAR+$^{\text{Emb}}$ &  17.89 &  21.09 &  29.76 &  33.59 &  36.69 &  37.31 &  40.51 \\
   & OSCAR+$^{\text{Ada}}$ &  18.52 &  23.94 &  31.45 &  34.60 &  37.26 &  37.97 &  40.60 \\
   & mBERT$^{\text{Ada}}$ &\bf19.77 &\bf31.99 &  34.49 &  36.26 & 39.15 &  39.81 &  40.88 \\
     \midrule    
\multirow{4}{*}{bn} & M3P &\bf17.59 &  17.33 &\bf26.94 &\bf31.09 &\bf34.58 &\bf35.27 &\bf37.96 \\
   & OSCAR+$^{\text{Emb}}$ &  13.35 &\bf17.40 &  21.67 &  26.61 &  31.94 &  32.78 &  36.97 \\
    & OSCAR+$^{\text{Ada}}$ &  13.96 &  15.60 &  22.35 &  27.20 &  31.25 &  31.81 &  35.45 \\
   & mBERT$^{\text{Ada}}$ &  13.38 &  11.33 &  23.10 &  26.55 &  31.60 &  32.26 &  34.18 \\
  \midrule
\multirow{4}{*}{ko} & M3P &  19.70 &\bf22.94 &\bf32.28 &\bf35.50 &\bf37.72 &\bf37.84 &\bf38.61 \\
   & OSCAR+$^{\text{Emb}}$ &  15.11 &  16.43 &  19.99 &  24.78 &  29.48 &  30.43 &  35.59 \\
   & OSCAR+$^{\text{Ada}}$ &  12.25 &  15.48 &  20.73 &  25.97 &  31.37 &  32.20 &  35.41 \\
  & mBERT$^{\text{Ada}}$ &\bf19.92 &  17.71 &  27.83 &  31.27 &  34.44 &  35.03 &  36.51 \\
      \midrule
\multirow{4}{*}{zh} & M3P &  19.66 &\bf27.76 &\bf36.15 &\bf38.21 &\bf40.48 &\bf40.53 &\bf42.55 \\
   & OSCAR+$^{\text{Emb}}$ &  12.66 &  14.77 &  19.17 &  22.13 &  27.97 &  29.08 &  33.24 \\
    & OSCAR+$^{\text{Ada}}$ &  13.20 &  15.12 &  19.67 &  22.74 &  26.81 &  28.19 &  31.69 \\
   & mBERT$^{\text{Ada}}$ &\bf26.16 &  23.47 &  32.93 &  35.82 &  38.22 &  37.89 &  39.57 \\
  \bottomrule
\end{tabular}
}
    \caption{Average accuracy of few-shot results, utilizing  different amounts of training data. The \textit{0} column presents the best zero-shot results. These models are used as initialization for the subsequent few-shot experiments. \textbf{Bold} numbers indicate the best scores.}
    \label{tab:fewshot_results}
\end{table}

\subsection{Few-Shot Cross-Lingual Transfer}

\begin{figure}[t!]

        \centering
        \includegraphics[width=\columnwidth]{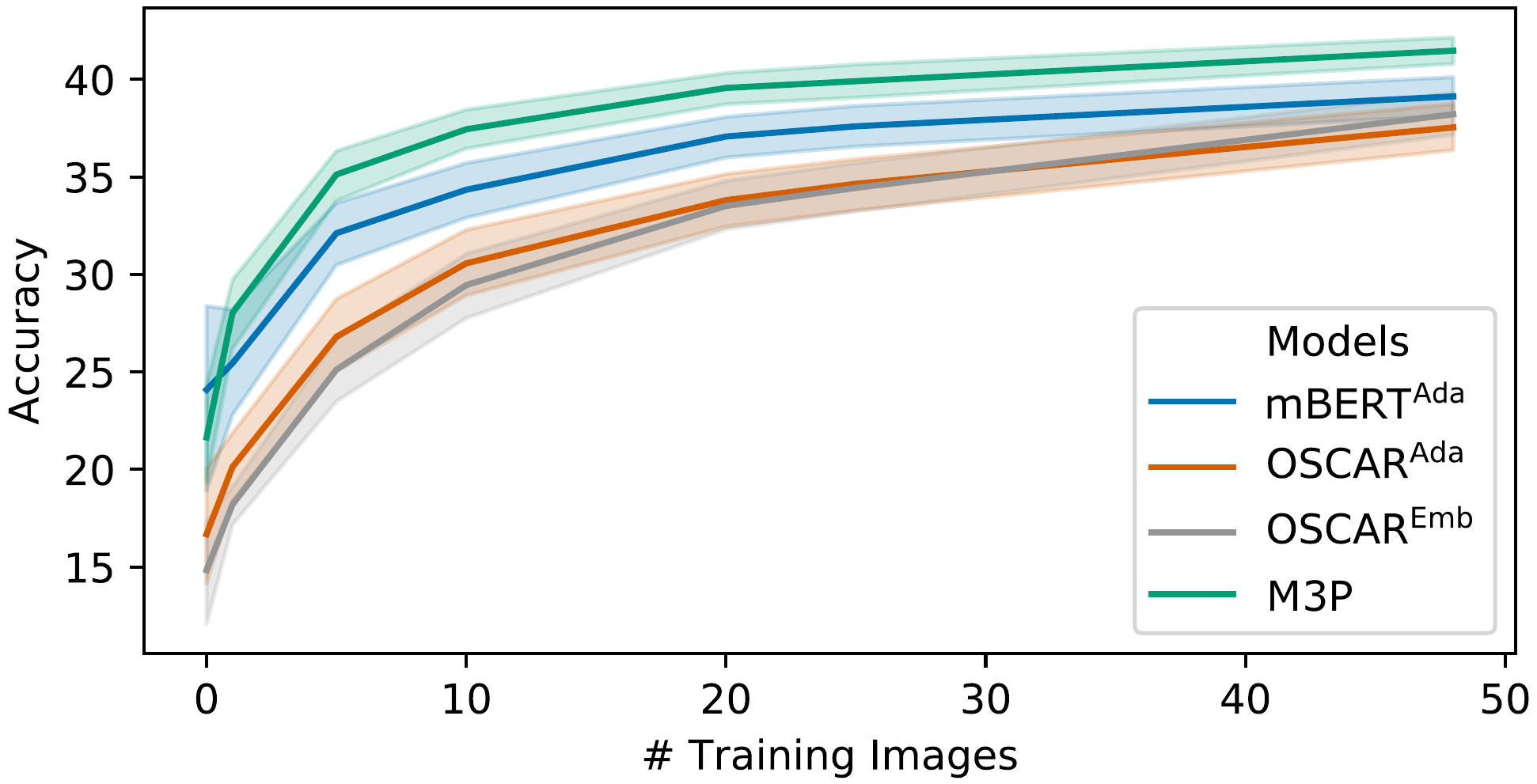}
    \caption{Few-shot accuracy with different training dataset sizes of the different approaches. Scores are averaged over all languages.}
    \label{fig:fewshot_average}
\end{figure}

The main results of few-shot experiments are provided in Table~\ref{tab:fewshot_results}, while the plot illustrating the impact of different amounts of training data is shown in Figure~\ref{fig:fewshot_average}.
One crucial finding is that, as expected, utilizing an increasing amount of data instances in the target language consistently improves accuracy for all methods. 
This culminates in an improvement of up to 20 accuracy points when specializing the model with only 48 images in the target language. This indicates that a small number of target-language examples supports the models in partially repairing its internal cross-lingual multimodal alignment.  
Interestingly, we find that with as little as 5 images, and their corresponding questions, M$^3$P begins to outperform mBERT$^\text{Ada}$---the best performing zero-shot model. 

We again analyze the impact of few-shot learning on accuracy across different structural question types, with the results depicted in Figure~\ref{fig:fewshotstructural}. The overall accuracy increases across all types compared to zero-shot scenarios (cf., Figure~\ref{fig:zerostructural}). However, the most pronounced gains are reported for \textit{query} and \textit{chose}-type questions, on which the model performed the worst in zero-shot setups. This implies the improved alignment between latent multimodal and multilingual representations, achieved via fine-tuning the model on a small amount of examples in the target language.

\subsection{Language Transfer}
We witness cross-lingual transfer capability patterns similar to those shown by previous work, where our models perform best on typologically close languages \cite{pires:2019,Lauscher:2020zerohero}. Our models transfer best to German (de) and Portuguese (pt), both being part of the Indo-European (IE) language family and also sharing the same script (Latin) with the source language English (en). We see a small drop in accuracy for Russian (ru), Indonesian (id), and Chinese (zh) and a larger drop in accuracy for Bengali (bn) and Korean (ko). All of these languages are typologically different to the source language and in most cases do not share the same script. These differences 
highlight the importance of language diversity in  cross-lingual transfer. Our benchmark thus  enables experimentation and  evaluation of multilingual multimodal models on a representative set of truly typologically diverse languages.

\section{Contemporary Work}
With the recent rise in interest in multilingual vision and language learning, contemporary work has already further analyzed and extended the proposed xGQA dataset. We provide a brief description and pointers to this work in what follows.

\vspace{1.6mm}
\noindent \textbf{Further Analysis.} 
\citet{Liu2022delvingdeeper} provide an extensive analysis of multilingual and multimodal models trained on cross-lingual visual question answering, and propose several approaches to mitigate the multilingual misalignment problem discussed in \S\ref{ss:zero_results}. Their results suggest that standard approaches
taken from text-only cross-lingual transfer scenarios \cite{pires:2019, Hu2020xtreme}  do not leverage the full multilingual capability of the pretrained models. Interestingly, they find that a deeper prediction head does not have any measurable impact on the model's performance in the source language, while at the same time it considerably improves zero-shot transfer results across all target languages.

\vspace{1.6mm}
\noindent \textbf{Translated Test Data.}
\citet{Bugliarello2022IGLUE} propose the first benchmark for transfer learning
across modalities, tasks, and languages, covering visual question answering,
cross-modal retrieval, grounded reasoning, and
grounded entailment tasks across 20 diverse languages. They extend the xGQA dataset by providing machine translated test-set questions and evaluate state-of-the-art monolingual multimodal models in a translate-test setup. In this setting, they achieve slightly better results. However, the performance remains to fall behind source language performance. The translate-test data can be found at \href{https://iglue-benchmark.github.io/}{iglue-benchmark.github.io}.

\section{Conclusion}
We have proposed xGQA, a first cross-lingual evaluation benchmark for the visual question answering task. xGQA extends the English GQA dataset with development and test data in 7 more typologically diverse languages, covering 5 different scripts. As additional baselines, we have further proposed new adapter-based methods to extend unimodal multilingual models to become multimodal and---vice-versa---monolingual multimodal models to become multilingual. Our results have indicated that 1) efficient adapter-based methods slightly outperform the 
pretrained multilingual multimodal model M$^3$P in zero-shot scenarios, but 2) the overall zero-shot cross-lingual transfer yields harsh accuracy drops compared to the English performance for all models in comparison. Further, accuracy can be partially recovered via few-shot learning, where small amounts of training data are available in the target language. However, the large gaps remain, suggesting the inherent complexity of the cross-lingual task despite it being extremely intuitive and easy to solve by (bilingual) humans.

We hope that our dataset and error analysis will motivate future work on this task and, more broadly, in the exciting emerging domain of multilingual multimodal representation learning.

\section*{Acknowledgments}

The Ubiquitous Knowledge Processing Lab acknowledges the financial support of the German Federal Ministry of Education and Research (BMBF) under the promotional reference 13N15897 (MISRIK), and the LOEWE initiative (Hesse, Germany) within the emergenCITY center.
Jan-Martin O. Steitz is supported by the LOEWE initiative (Hesse, Germany) within the emergenCITY center. The work of Ivan Vuli\'{c} is supported by a Huawei research donation and the ERC PoC Grant MultiConvAI: Enabling Multilingual Conversational AI (no. 957356). Stefan Roth is additionally supported by the European Research Council (ERC) under the European Union’s Horizon 2020 research and innovation programme (grant agreement No.~866008). 

We thank Leonardo F. R. Ribeiro, Ji-Ung Lee, and Chen  Liu for insightful feedback and suggestions on a draft of this paper.
 
\bibliography{anthology,custom}
\bibliographystyle{acl_natbib}


\appendix

\section{Appendix}
\label{sec:appendix}
\begin{figure*}[t!]
    \centering
    \begin{subfigure}[b]{0.195\textwidth}
        \centering
        \includegraphics[width=0.9\columnwidth]{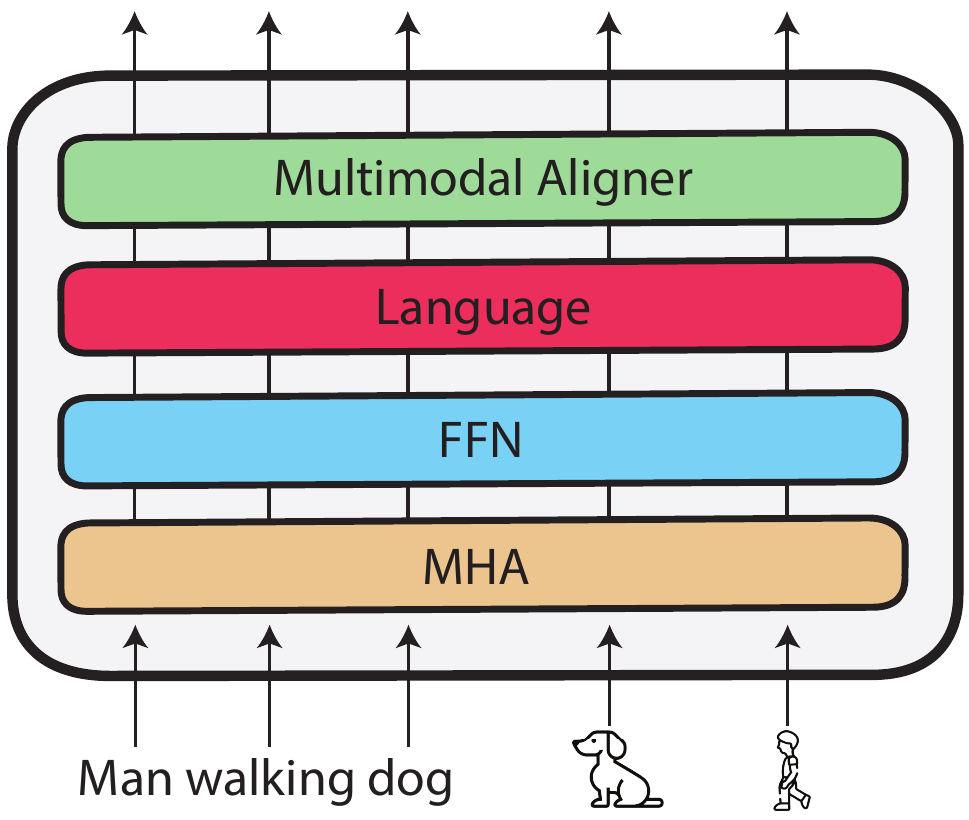}
        \caption{Setting 1}
        \label{fig:ada_setting2}
    \end{subfigure}
    \begin{subfigure}[b]{0.195\textwidth}
        \centering
        \includegraphics[width=0.9\columnwidth]{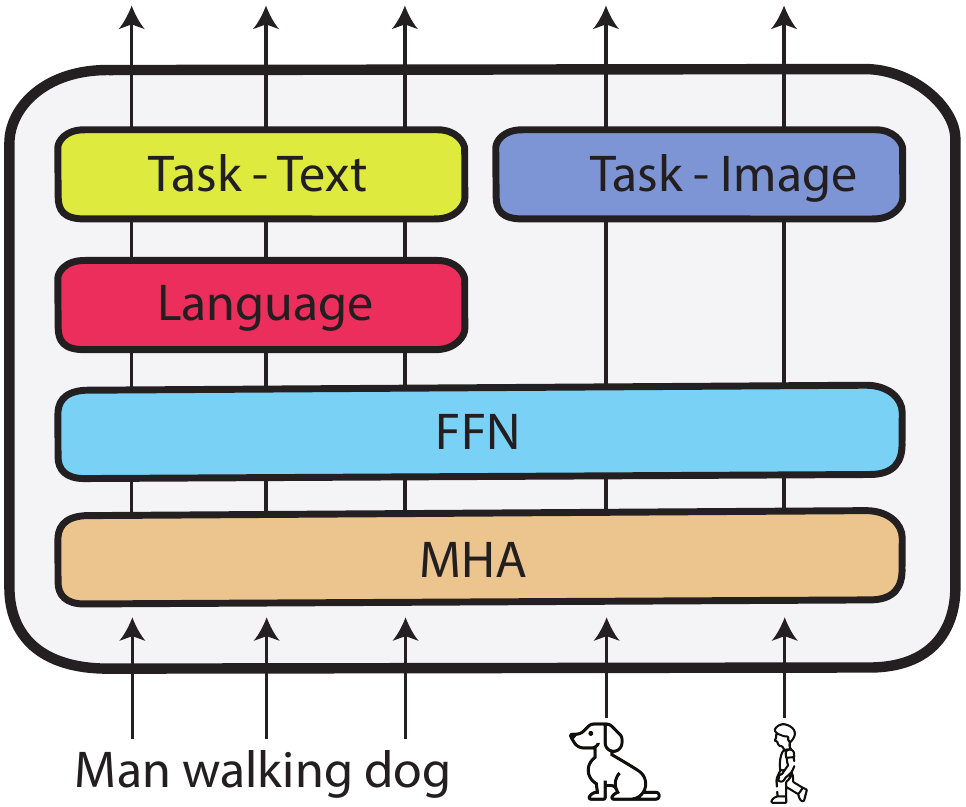}
        \caption{Setting 2}
        \label{fig:ada_setting3}
    \end{subfigure}
    \begin{subfigure}[b]{0.195\textwidth}
        \centering
        \includegraphics[width=0.9\columnwidth]{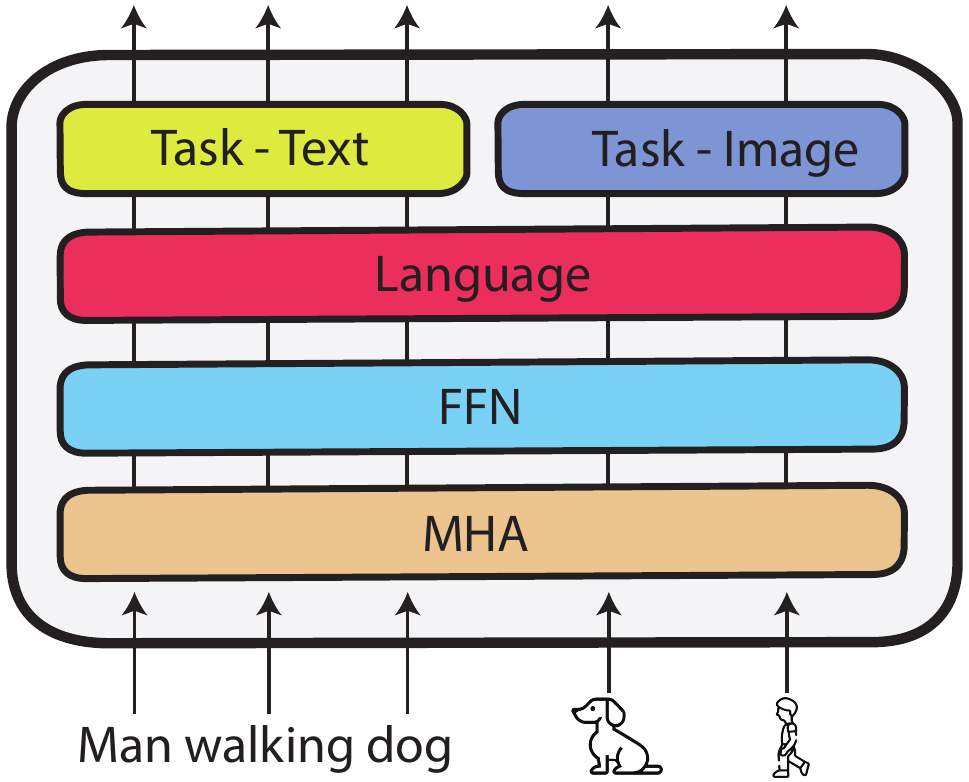}
        \caption{Setting 3}
        \label{fig:ada_setting4} 
    \end{subfigure}
    \begin{subfigure}[b]{0.195\textwidth}
        \centering
        \includegraphics[width=0.9\columnwidth]{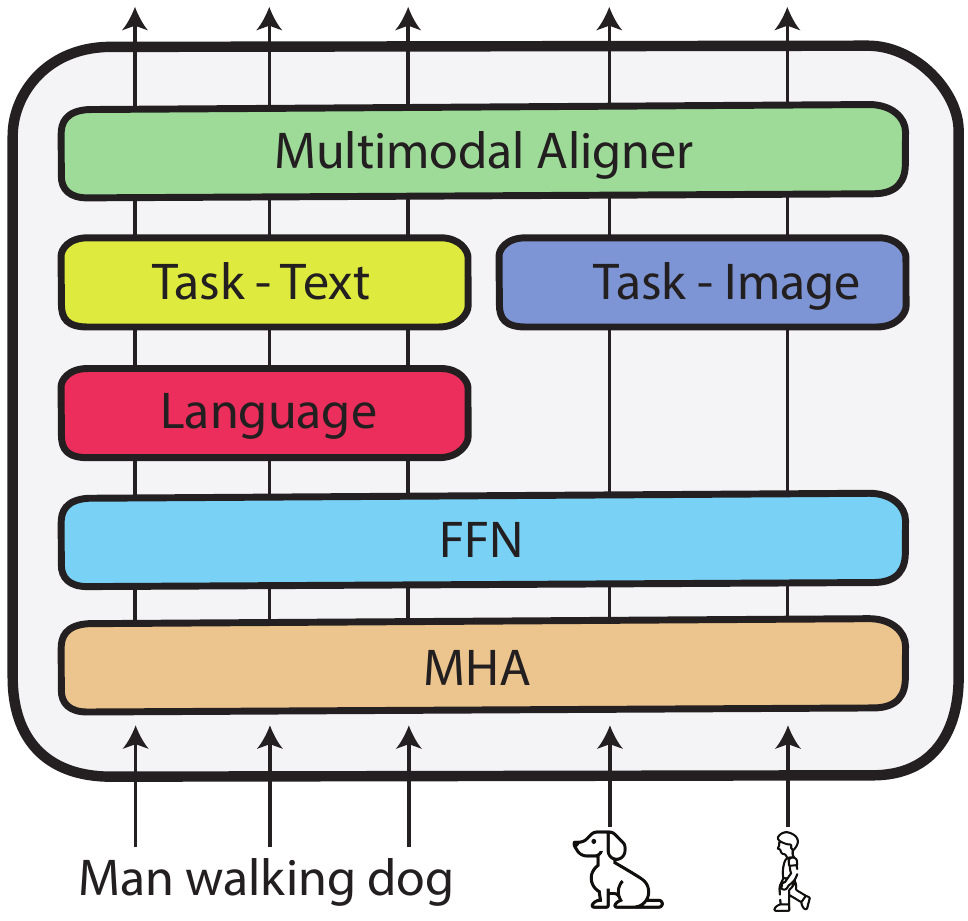}
        \caption{Setting 4}
        \label{fig:ada_setting5} 
    \end{subfigure}
    \begin{subfigure}[b]{0.195\textwidth}
        \centering
        \includegraphics[width=0.9\columnwidth]{img/archs/xgqa_arch_1.pdf}
        \caption{Setting 5}
        \label{fig:ada_setting5} 
    \end{subfigure}
    \caption{The different multimodal multilingual adapter architectures we experimented with. The best performing architecture was setting 5, for which we present results in the main paper. }
    \label{fig:alternativ_ada_settings}
\end{figure*}

We experiment with different multimodal adapter architectures as illustrated in Figure~\ref{fig:alternativ_ada_settings}. In initial experiments we find that splitting the modalities (settings 2-5) outperforms a joint adapter (setting 1). However, a joint "alignment" architectures (settings 4-5) outperform settings where we only use modality-specific adapters (settings 2-3). We more thoroughly investigate settings 4-5 and report scores in Table~\ref{tab:different_adapters}. Interestingly, we find that when only using the language adapter for the textual inputs, cross-lingual accuracy drops for both OSCAR+ and mBERT; The difference is more pronounced for OSCAR+. We speculate that this is due to a latent misalignment of the representation spaces, partly due to the residual connection. Due to the better performance of setting 5 on average, we have reported scores of this architecture in the main paper (as illustrated in Figure~\ref{fig:adapter_architecture}).

\begin{table*}[t!]
\centering
\resizebox{1.0\textwidth}{!}{%
\begin{tabular}{llc:lllllll|l}
\toprule
model &   Setting &   en &     de &     pt &     ru &     id &     bn &     ko &     zh &   mean   \\
\midrule
 OSCAR+$^{\text{Ada}}$ &   \multicolumn{1}{c}{4} &  60.21 &  18.60 &  25.48 &   8.22 &  17.79 &  10.47 &   9.97 &  12.54 &  14.72 \\
OSCAR+$^{\text{Ada}}$ &   \multicolumn{1}{c}{5} & \bf  60.30  & \bf 18.91  & \bf 27.02  & \bf 17.50  & \bf 18.77  &  \bf 15.42  & \bf 15.28  & \bf 14.96  & \bf 18.27 \\
\hdashline
mBERT$^{\text{Ada}}$  &   \multicolumn{1}{c}{4} & \bf 57.83 &  27.86 &  28.88 &  22.87 & \bf  20.86 &  14.74 &  18.30 &  24.39 &  22.56\\
mBERT$^{\text{Ada}}$ &   \multicolumn{1}{c}{5} &   56.25  & \bf   29.76  & \bf  30.37 & \bf  24.42 &  19.15  &  \bf 15.12  & \bf  19.09  & \bf  24.86  &  \bf 23.25 \\
\bottomrule
\end{tabular}
}
\caption{Zero-shot transfer results on xGQA for the different adapter architecture settings (as illustrated in Figure~\ref{fig:alternativ_ada_settings}) when transferring from English GQA. Average accuracy is reported. Best results for each language and model type are  highlighted in \textbf{bold}; \textit{mean} scores are not averaged over the source language (English).}
\label{tab:different_adapters}
\end{table*}

\end{document}